\documentclass[letterpaper]{article} 
\usepackage[]{aaai24}  
\usepackage{times}  
\usepackage{helvet}  
\usepackage{courier}  
\usepackage[hyphens]{url}  
\usepackage{graphicx} 
\usepackage[numbers]{natbib}
\usepackage{caption} 
\usepackage{booktabs}
\usepackage{algorithm}
\usepackage{algorithmic}

\usepackage{newfloat}
\usepackage{listings}
\DeclareCaptionStyle{ruled}{labelfont=normalfont,labelsep=colon,strut=off} 
\floatstyle{ruled}
\newfloat{listing}{tb}{lst}{}
\floatname{listing}{Listing}
\title{Text Sanitization Beyond Specific Domains: \\
Zero-Shot Redaction \& Substitution with Large Language Models}

\author{%
  Federico Albanese \textsuperscript{\rm 1}, 
  Daniel Ciolek \textsuperscript{\rm 2}, 
  Nicolas D'Ippolito \textsuperscript{\rm 3}, 
}
\affiliations{
    \textsuperscript{\rm 1} Instituto en Ciencias de la Computación, CONICET - Universidad de Buenos Aires, Argentina\\
    \textsuperscript{\rm 2} Universidad Nacional de Quilmes, Argentina\\
    \textsuperscript{\rm 3} ASAPP\\

    falbanese@dc.uba.ar.com
}

\begin{document}

\maketitle

\begin{abstract}
In the context of information systems, text sanitization techniques are used to identify and remove sensitive data to comply with security and regulatory requirements. Even though many methods for privacy preservation have been proposed, most of them are focused on the detection of entities from specific domains (e.g., credit card numbers, social security numbers), lacking generality and requiring customization for each desirable domain. Moreover, removing words is, in general, a drastic measure, as it can degrade text coherence and contextual information. Less severe measures include substituting a word for a safe alternative, yet it can be challenging to automatically find meaningful substitutions. We present a zero-shot text sanitization technique that detects and substitutes potentially sensitive information using Large Language Models. Our evaluation shows that our method excels at protecting privacy while maintaining text coherence and contextual information, preserving data utility for downstream tasks.
\end{abstract}

\section{Introduction}

The ever-increasing volume of data stored in software systems raises concerns about the potential exposure of sensitive information, such as Personally Identifiable Information (PII) or other confidential data (e.g., medical diagnoses) \cite{lison2021anonymisation}. To safeguard individuals' privacy, processes of data sanitization have emerged as a crucial practice to redact sensitive information in compliance with existing regulations, such as the General Data Protection Regulation (GDPR) \cite{goddard2017eu}, Health Insurance Portability and Accountability Act (HIPAA) \cite{nosowsky2006health}, and California Consumer Privacy Act (CCPA) \cite{goldman2020introduction} in the US and EU regions.

The importance of sensitive data redaction cannot be overstated, as it plays a key role in ensuring privacy when publishing documents, interacting with third-party APIs, and training Artificial Intelligence (AI) models \cite{vasudevan2014review}. As AI continues to revolutionize various domains, it heavily relies on vast amounts of data for training. The sheer volume of data makes it essential to adopt automatic redaction systems \cite{sanchez2013automatic}. However, the nature of textual data makes the accurate detection and redaction of sensitive information challenging.

Hereinafter we propose a technique for text sanitization in dialogue systems (i.e., unstructured conversation text lacking explicitly identifiable attributes). Our approach uses a Large Language Model (LLM) to first estimate the probability of a word in the textual context, then it redacts words with a probability below a specified privacy threshold, and finally it substitutes redacted words with semantically close alternatives by leveraging LLM word embeddings.

Consider the example dialogue text "My name is John Smith." containing two terms considered PII, the names "John" and "Smith". Using a LLM, we can estimate the probability of each word in the text, for instance “My”: $0.03$, “name”: $0.07$, “is”: $0.06$, “John”: $0.004$, “Smith”: $0.001$. Words falling beneath a predetermined “privacy threshold” $p$ are redacted and substituted. In this example, if we use $p = 0.01$, both "John" and "Smith" satisfy this criterion. For these terms, alternative substitutions can be found also using the LLM, for instance, replacing "John" with "David" and "Smith" with "Williams". The output of the model is the sanitized text "My name is David Williams". This output preserves privacy, coherence, and meaning, rendering it suitable for dissemination and other tasks. 

Having a configurable privacy threshold allows our technique to achieve different trade-offs between safety guarantees and data utility. In contrast, rule-based and Named Entity Recognition (NER) approaches are restricted to a specific set of rules and entities. Moreover, even with high privacy thresholds, applying substitution allows our technique to keep enough information to preserve the performance of downstream tasks, in contrast to plain redaction.

The generalization capabilities of LLM makes them well-suited for handling unstructured text data. Our method leverages pre-trained LLM, allowing us to perform zero-shot text sanitization, reducing the time to value in real world scenarios. Considering that this approach is compatible with multilingual language models, our technique is also useful for text sanitization in multiple languages with no additional cost.

The main contribution of this work is a Zero-Short Text Sanitization (ZSTS) system with the following characteristics:

\begin{itemize}

\item {\bf Zero-shot}: By relying on a pre-trained LLM, the model does not require training, making it domain independent, and significantly reducing time-to-value during deployment in real-world applications.

\item {\bf Privacy protection}: By using a high privacy threshold, the model can achieve high performance in the detection of potentially sensitive terms. 

\item {\bf Semantic preservation}: By leveraging LLMs word embeddings we can substitute words in such a way that maintains text coherence and contextual information, preserving data utility and performance in downstream tasks.

\item {\bf Multilingual}: By using an LLM pre-trained with text from different languages, our approach can be used to redact text in a multilingual setting.

\end{itemize}

\section{Background}


The practice of document sanitization involves the removal of sensitive terms from a document \cite{lison2021anonymisation}. Sensitive terms are such that, due to ther specificity, provide more information relative to non-sensitive terms. In the hypothetical context of an audience (or attacker) possessing basic information about a corpus of documents, the optimal approach to document sanitization should remove the terms that would increase the audience's existing knowledge. Consequently, the key of detecting sensitive information lies in evaluating the information that each term conveys, followed by the removal of the terms which exceed the assumed knowledge of the audience.

We can see the occurrence of a word in a text as a random event following some (unknown) probability distribution. In information theory, the Information Content (IC) of an event $e$ is a quantity derived from the probability of $e$ occurring from a random variable.
Given event $e$ with probability $P_e$ its IC is defined as $\mathit{IC}(e) = -\log(P_e)$. That is, an event with probability $1$, is considered to contain $0$ information, and as we consider lower probabilities the IC increases monotonically. Thus, removing terms above a certain IC threshold (or below an analogous probability threshold), would decrease a document’s overall IC.

An LLM outputs a probability distribution over sequences of words \cite{vaswani2017attention}.
Terms with low probability have higher IC, and hence are more likely to represent sensitive information. 
Thus, we can use an LLM to estimate the probability of each word in a document, and filter words with an estimated probability below a privacy threshold $p$.

In addition to outputting a distribution over sequences of words, LLM are also capable of computing embeddings for individual words, sentences, or entire documents \cite{devlin2019bert}. An embedding is a representation of the text in a high-dimensional vector space, where the semantic relationships between words or sentences correlate with the relationships in the vector space. Thus, we can compare the similarity between embeddings relying on metrics such as cosine similarity.

\section{Redaction and Substitution}

Our ZSTS system consists of three distinct phases:

\begin{enumerate}
\item {\bf Preprocessing}: The document is tokenized, splitting it into words.
Importantly, the choice of the tokenization method needs to be consistent with the LLM's, which may require to tokenize some words in sub-word units.
To reduce the total vocabulary size, all tokens are turned into lowercase.

\item {\bf Privacy-threshold filtering (aka. p-filtering)}: To calculate the probability of a word within a document, we utilize a masking technique where the word is replaced with a masking token, and an LLM estimates the probability distribution for the masked word.
In the case that a word is composed of multiple sub-word tokens, we apply this process iteratively, first masking the entire word, and then gradually unmasking the beginning of the word while keeping the trailing tokens masked.
The final probability of the multi-token word is aggregated by taking the product of the probabilities of its constituent tokens.

\item {\bf Substitution}: For numeric tokens, we generate substitutions with randomly generated numbers of the same length.
For non-numeric tokens we use the following procedure (for a choice of hyper-parameters $n$, $k$, and $s$):
\begin{enumerate}
\item Select the top $n$ most probable candidates for the masked word, based on the probability distribution calculated in the previous step;
\item Sort the candidates by similarity to the original word (word distance is determined by computing the cosine distance between LLM word embeddings); and
\item Pick a candidate at random from the top $k$ most similar words within a distance radius $s$, and if there are less than $k$ words within distance $s$, remove the term instead.
\end{enumerate}
The random choice prevents deterministically recovering the original terms from their substitutions, adding an extra layer of protection against adversarial attacks.
Additionally, we keep a substitution table to guarantee consistent replacement throughout the document.
This ensures that if a sensitive word reappears in the text, it is replaced by the same designated substitution.
\end{enumerate}


Redaction and substitution can be performed either separately or simultaneously, that is, where the estimation of the probability of a word and its potential replacement is computed in a single invocation of the LLM.
On the one hand, this approach reduces the computational cost.
On the other hand, performing these steps independently allows using different models for redaction and substitution.

This framework is compatible with various LLM. In this paper, we report our findings using BERT (Bidirectional Encoder Representations from Transformers) \cite{devlin2019bert}, which has been influential in many benchmark tasks and it is widely used in the literature. The bidirectional context modeling inherent in BERT provides a good capability to comprehend context both prior to and after the masked token. This allows for a better probability estimation compared to models like GPT2 \cite{radford2019language}, which focus exclusively on the preceding context. Since BERT is pre-trained with public data, it is not exposed to the sensitive data under redaction. Also, pre-trained LLM like BERT enables local execution without the need to share sensitive data with third parties during the redaction and substitution process, in contrast to utilizing an LLM through an external API. 

\section{Evaluation}

\subsection{Datasets}

For the evaluation we use the Action-Based Conversation Dataset (ABCD) \cite{chen2021action}, which is a good representative of dialogue systems. ABCD contains over 10K human-to-human dialogues with 55 distinct user intents and more than 140K utterances. Importantly, the dataset contains metadata identifying distinctive slot values and subset of these values corresponds to (synthetic) PII. For our evaluation we consider as PII the following categories in the metadata: \textit{customer name}, \textit{username}, \textit{email}, \textit{phone number}, \textit{account id}, \textit{order id}, \textit{street address}, and \textit{zip code}.

\subsection{Baselines}

We consider several established models for comparison with our proposed redaction model. These baseline models include:

\begin{itemize}

\item Microsoft Presidio \cite{MSFT}: Microsoft Presidio is a widely adopted redaction library that employs rule-based methods and NER to identify and redact sensitive information.

\item Google Data Loss Prevention \cite{Google}: Google DLP is the de-identification service for masking, deleting or obscuring sensitive data such as PII, which leverages various techniques including pattern matching, checksums, machine-learning, and context analysis. 
\end{itemize}

We chose Google DLP and Microsoft Presidio as our baselines due to their widespread usage in the literature, ensuring a standard baseline for our research \cite{hassan2021utility, lison2021anonymisation, mansfield2022behind, zhou2023privacy}.

\subsection{Metrics}

We can understand redaction as the task of classifying unsafe terms. Thus, we can rely on standard classification metrics:
\[
\mathit{Precision} = \frac{\mathit{TruePositives}}{\mathit{TruePositives} + \mathit{FalsePositives}}
\]
\[
\mathit{Recall} = \frac{\mathit{TruePositives}}{\mathit{TruePositives} + \mathit{FalseNegatives}}
\]

Where:
\begin{itemize}

\item $\mathit{TruePositives}$, is the number of unsafe terms properly predicted as unsafe;

\item $\mathit{TrueNegatives}$, is the number of safe terms properly predicted as safe;

\item $\mathit{FalsePositives}$, is the number of safe terms wrongly predicted as unsafe; and

\item $\mathit{FalseNegatives}$, is the number of unsafe terms wrongly predicted as safe.

\end{itemize}

In this setting Recall indicates the proportion of unsafe terms properly redacted.
That is, high Recall means most unsafe terms are properly redacted.
Whereas Precision indicates the proportion of unsafe terms redacted over all redacted terms.
That is, a low Precision is an indication of over-redaction. 

As it is customary, we can combine both metrics with the F1-score, defined as the harmonic mean between precision and recall:
\[
\mathit{F1} = 2 \frac{\mathit{Precision} \cdot \mathit{Recall}}{\mathit{Precision} + \mathit{Recall}}
\]

\subsection{Experiments}

We report on two distinct experiments, the first focuses on redaction quality, and the second on the impact of the substitution step on downstream tasks.

\subsubsection{Redaction Quality}

In our experiment we process ABCD texts utterance by utterance (as they would have reached a backend if running a realtime chat system). This strategy allows the LLM to estimate accurate probabilities for sensitive terms even when they are recurrently mentioned through a conversation (e.g. names of the interlocutors). Additionally, the preceding redacted utterance is provided as context, which is particularly valuable for contextualizing an utterance (e.g., a response to a question). Finally, using ABCD metadata, we automatically determine how many sensitive terms survive after applying redaction.

In Table \ref{table1_2}, we detail the Recall, Precision, and F1 scores for both baseline models and our ZSTS system, which employs BERT for estimating word probabilities and redacting sensitive words.

\begin{table*}[h!]
\centering
\begin{tabular}{lccc}
    \toprule
    Model     & Recall & Precision & F1\\
    \midrule
    Microsoft Presidio & $0.56$  & $0.30$ & $0.39$  \\
    Google DLP    & $0.65$  & $0.66$ & $0.66$  \\
    \midrule
    ZSTS ($p{=}1E^{-25}$) & $0.06$  & $0.81$  & $0.10$     \\
    ZSTS ($p{=}1E^{-15}$) & $0.26$  & $0.80$  & $0.40$     \\
    ZSTS ($p{=}1E^{-5}$) & $0.54$  & $0.29$  & $0.37$      \\
    ZSTS ($p{=}1E^{-4}$) & $0.68$  & $0.21$  & $0.32$      \\
    ZSTS ($p{=}1E^{-3}$) & $0.84$  & $0.15$  & $0.26$      \\
    ZSTS ($p{=}1E^{-2}$) & $0.94$  & $0.10$  & $0.19$      \\
    ZSTS ($p{=}1E^{-1}$) & $0.98$  & $0.06$  & $0.12$      \\
    \bottomrule
\end{tabular}
\caption{Detailed redactions results for baseline models and ZSTS ($\textit{LLM}{=}\mbox{BERT}$, $n{=}50$, $s{=}2$, $k{=}1$).}
\label{table1_2}
\end{table*}

The proposed model achieves its maximum F1 at $p {=} 1E^{-15}$.
Interestingly, F1 scores for $p {=} 1E^{-25}$ and $p {=} 1E^{-1}$ are similar, showcasing high Recall and low Precision, and vice versa.
There is a trade-off between Recall and Precision.
The values of $p$ that allow achieving a high Recall (i.e., redact most sensitive terms), entail a low Precision (i.e., over-redaction). Varying $p$, users can flexibly determine the level of redaction most appropriate for their use case.

Our approach offers users the advantage of achieving high Recall, which is a crucial aspect for real-world applications. However, high Recall comes at the cost of low Precision. When applying only the redaction step, low precision implies a substantial loss of information, potentially compromising performance in downstream tasks. To address this challenge, we introduce the substitution step. This additional step is designed to mitigate information loss resulting from excessive redaction, striking a balance between privacy and text utility, and enhancing the technique's effectiveness in practical applications.

\subsubsection{Substitutions Quality}

Achieving low Precision while substituting words instead of simply redacting them, raises the question of how much information is actually preserved by the substitution step. We aim to measure this by analyzing the effect of substitution in different downstream tasks.

\begin{enumerate}

\item {\bf Sentiment Analysis}: We determine if the sanitization process changes the original text's sentiment using sentiment analysis with the VADER library \cite{hutto2014vader}. By comparing VADER scores (ranging from -1 to 1) of the original and sanitized text, sentiment is classified as positive ($>0.05$), negative ($<-0.05$), or neutral. Our model's success in keeping sentiment labels after substitution is evaluated by accuracy.

\item {\bf Topic Embedding}: To evaluate the effect of sanitization on a text's perceived topic, each conversation in ABCD is categorized using BertTopic \cite{grootendorst2022bertopic}, a model that uses Bert and unsupervised clustering to estimate the count and probabilities of topics in the text (i.e., BerTopic finds 33 topics in the ABCD). Then, the average cosine distance between the topic vector of the original and sanitized texts is computed. A distance of 0 suggests identical topic decomposition, while 1 represents disjoint topics. This metric indicates whether the main topic remains unchanged by substitutions.

\item {\bf Question Answering (Q\&A)}: We examine the sanitization impact on Q\&A using GPT3.5 \cite{brown2020language}. The test includes questions around mentioned entities (names, usernames, emails), expressions of dissatisfaction, and whether conversations were left incomplete. We compare GPT3.5's responses to the same prompt from both raw and sanitized conversations. Using the substitution table in reverse, we recover the original potentially sensitive words by replacing back the occurrences of substitutions, and we verify if the answer matches the one provided for the original (unredacted) conversation. The model's performance is measured by an accuracy score, with 1 being a perfect match to the original, unredacted conversation answer, and 0 being no matches. Conversations lacking entities, lead to significantly easier matches. Thus, we report distinct accuracy for scenarios with retrievable entities or where the original answer was true. For this experiment we use a random subset of 50 ABCD conversations due to computational constraints.

\end{enumerate}

\begin{table*}[t]
\centering
\begin{tabular}{lllllll}
\toprule
        & Privacy & Sentiment & Topic distance & Q\&A &  Q\&A true \\
    Method     &  [Recall] & [Accuracy] & [mean$\pm$std] &  [Accuracy] &   [Accuracy] \\
    \midrule
    Presidio & $0.56$  & $0.999$ & $0.001 \pm 0.001$ & $0.61$  & $0.52$ \\
    DLP & $0.65$  & $1$ & $0.0003 \pm 0.0006$ & $0.52$  & $0.36$ \\
    ZSTS Redaction & $0.98$  & $0.997$ & $0.022 \pm 0.021$ & $0.57$  & $0.26$ \\
    ZSTS Redaction \& Substitution & $0.98$  & $0.993$ & $0.023 \pm 0.021$ & $0.82$  & $0.75$ \\
    \bottomrule
\end{tabular}
\caption{
Results of the effects of text sanitization on privacy and downstream tasks for baseline models and ZSTS ($\textit{LLM} {=} \mbox{BERT}$, $p{=}1E^{-1}$, $n{=}50$, $s{=}2$, $k{=}1$).}
\label{table3}
\end{table*}

The results in Table \ref{table3} show that:

\begin{itemize}

\item As it is to be expected, techniques that redact less (i.e., achieve a lower Recall), tend to have a smaller impact on downstream tasks.

\item Despite achieving a high Recall, the effects of ZSTS on Sentiment Analysis and Topic Embedding are very small.

\item Redaction alone (i.e., no substitution) significantly reduces data utility for Q\&A for both ZSTS and the baseline models.

\item Doing redaction and substitution with ZSTS achieves high Recall, while preserving accuracy for Q\&A and introducing a negligible degradation on Sentiment Analysis and Topic Embedding. Importantly, when an answer is true, in general an entity needs to be recovered from the text (e.g., “Was there a name mentioned?” $\rightarrow$ “Yes, David”). In such cases the advantage of applying substitution becomes preponderant.

\end{itemize}

In summary, ZSTS effectively protects privacy while also preserving text utility, in contrast to the baselines.

\section{Related work}

Automated text anonymization has been a topic of much interest.
In \cite{lison2021anonymisation} a survey on key concepts and current approaches to automated text anonymization is provided.
However, most previous works have adopted rule-based and NER approaches, which are restricted to a predefined set of rules and entities.
For example, in \cite{davidson2021improved} the authors describes a NER approach suitable for detecting entities that need to be redacted.
Similarly, \cite{mehta2019towards} adopts a two-phase Conditional Random Field for NER to represent and anonymize unstructured data.
However, NER techniques only detect a fixed set of entity types and have classifiers that need to be trained, requiring a large amount of manually tagged data \cite{hassan2021utility}.
Our approach, on the other hand, leverages a pre-trained LLM to estimate the probability of words in a textual context, enabling a zero-shot detection beyond predefined entities.

A related method for text sanitization is proposed in \cite{papadopoulou2022neural}, taking an approach to protect privacy by evaluating and masking (both direct and indirect) personal identifiers in the text.
Their technique provides explicit measures of reidentification risk, thus allowing a fine-tuned control over the balance between privacy protection and data utility.
Our approach aligns with them on adopting a BERT language model to estimate term probabilities.
However, our work also utilizes BERT to generate potential substitutions for the sanitization process, increasing data utility without sacrificing privacy.

Most previous approaches \cite{braathen2021creating, dernoncourt2017identification, pilan2022text, papadopoulou2022neural} replace the detected text spans by default strings or a black box.
However, some sanitization techniques choose to use substitute methods to preserve the contextual meaning and coherence of the original text.
A general-purpose sanitization method exploiting knowledge bases to compute term frequency for sensitive term substitution is proposed in \cite{sanchez2013automatic}.
Privacy-aware back-translation methods for sensitive attribute obfuscation is explored in \cite{xu2019privacy}.
An approach to generate possible replacements using a combination of heuristic rules and an ontology derived from Wikidata is presented in \cite{olstad2023generation}.
However, these methods are reliant on additional data sources for term substitutions.
Our approach uniquely leverages the semantic word embeddings of a pre-trained LLM to find alternative words ensuring the semantic coherence in sanitized texts.

Several studies have addressed the trade-off between safety and data utility concerning privacy threshold.
For instance, the ERASE framework \cite{chakaravarthy2008efficient} uses a property called k-safety.
Similarly, a “t-plausibility” model is proposed in \cite{anandan2012t}.
Others have explored the model of “K-confusability” \cite{cumby2011machine}, l-diversity \cite{machanavajjhala2007diversity} and probabilistic k-anonymity \cite{senavirathne2021systematic}, each casting a different light on preserving privacy while maintaining data utility.
Our work offers a different approach based on a privacy threshold considering the information content of a word in the textual context.

Preserving the performance in downstream tasks is an essential requirement that several studies have acknowledged by evaluating the performance of ML algorithms on redacted datasets \cite{senavirathne2020role, malle2017disturb, wimmer2014comparison}.
However, the primary focus in these works has been on structured datasets.
Our work enables the study of the preservation of performance in downstream tasks when subjected to unstructured text data.

\section{Conclusions}

We present ZSTS, a novel technique for text sanitization, particularly developed for unstructured text data. Our technique has several significant features, including zero-shot (i.e., no training required), privacy safeguarding, semantic preservation, and multilingual compatibility. Utilizing a pre-trained LLM, our approach estimates the probability of words in a given textual context. Words falling below a specified privacy threshold are redacted and substituted with semantically close alternatives that maintain the coherence and meaning of the original text.

Our evaluation shows that our method achieves a good tradeoff between privacy protection (i.e., high recall) and data utility preservation (i.e., good performance on downstream tasks), demonstrating its ability to redact and substitute sensitive information without unnecessary over-redaction. 

Our findings suggest notable improvements over the baseline approaches, especially in terms of data utility. For tasks such as Sentiment Analysis and Topic Embedding, our sanitization technique led to a minimal degradation in results while achieving a higher Recall. For Q\&A tasks, redaction combined with substitution allowed achieving a higher accuracy than the baselines.

The main drawback of our approach is that it incurs a high computational cost due to invoking an LLM for each word in the document to redact. As future work we plan to explore the adequacy of smaller language models to be used instead of LLM.
Additionally, our substitution strategy for numeric tokens, based on random sampling, fails to preserve the algebraic relations between numbers.
Another avenue for future work is to employ LLM to detect related numbers and generate alternatives that preserve such relations.
We evaluate our model using ABCD, a dialogue dataset. Future work would validate the proposed model in more diverse scenarios.

In summary, our approach demonstrates that it is possible to achieve efficient text sanitization, simultaneously protecting privacy and preserving data utility. These promising results underline the potential of this methodology for real-world applications.


\end{document}